\documentclass[letterpaper, 10 pt, conference]{ieeeconf}  
\usepackage{graphicx}
\usepackage{subcaption}
\usepackage{wrapfig}

\IEEEoverridecommandlockouts                              
\title{\LARGE \bf
ViZDoom: DRQN with Prioritized Experience Replay, Double-Q Learning, \& Snapshot Ensembling*
}

\author{\Large Christopher Schulze, Marcus Schulze
\\
\{cschulze7, mschulze765\}@gmail.com
\thanks{*This work was not supported by any organization}
}

\begin{document}

\maketitle
\thispagestyle{empty}
\pagestyle{empty}

\begin{abstract}

ViZDoom is a robust, first-person shooter reinforcement learning environment, characterized by a significant degree of latent state information. In this paper, double-Q learning and prioritized experience replay methods are tested under a certain ViZDoom combat scenario using a competitive deep recurrent Q-network (DRQN) architecture. In addition, an ensembling technique known as snapshot ensembling is employed using a specific annealed learning rate to observe differences in ensembling efficacy under these two methods. Annealed learning rates are important in general to the training of deep neural network models, as they shake up the status-quo and counter a model's tending towards local optima. While both variants show performance exceeding those of built-in AI agents of the game, the known stabilizing effects of double-Q learning are illustrated, and priority experience replay is again validated in its usefulness by showing immediate results early on in agent development, with the caveat that value overestimation is accelerated in this case. In addition, some unique behaviors are observed to develop for priority experience replay (PER) and double-Q (DDQ) variants, and snapshot ensembling of both PER and DDQ proves a valuable method for improving performance of the ViZDoom Marine.

\end{abstract}

\section{INTRODUCTION}

Increasingly, deep reinforcement learning (DRL) is the topic of great discussion in the artificial intelligence community . For reinforcement learning (RL) experts, RL has always shown great promise as a robust construct for solving task oriented problems. Recently, the DRL variant has gained significant public attention. But to be clear, much remains to be done at large in the field concerning high state-action dimensional problems that approximate well some real world problems of interest. From board games like Go in recent engagements with agents from Deep Mind to very high state-action dimensional games like those from the real time strategy (RTS) and first person shooter (FPS) video game genres, DRL has established itself, presently, as a prime construct for solving highly complex problems without direct supervision for most, if not all, of its training.

	Given the success, much work has been done recently to improve upon models of DRL, including that of the Deep Q-Network (DQN). Priority experience replay (PER) and Double-Q learning (DDQ) are two such methods that can be used to improve a DQN agent's rate of improvement or degree of learning stability, respectively, during training. However, to the authors knowledge, these two methods have yet to be tested under the ViZDoom [8] environment - a setting characterized by a higher degree of latent information relative to Atari and other popular RL environments - in the context of a specific, effective deep recurrent learning architecture. Replicating the DRQN structure in the paper from Lample et al.[5], the authors in this paper test the benefits offered by PER and DDQ methods under an efficient ensembling method.

\begin{figure}[h!]
  \centering
  \includegraphics[width=0.5\textwidth]{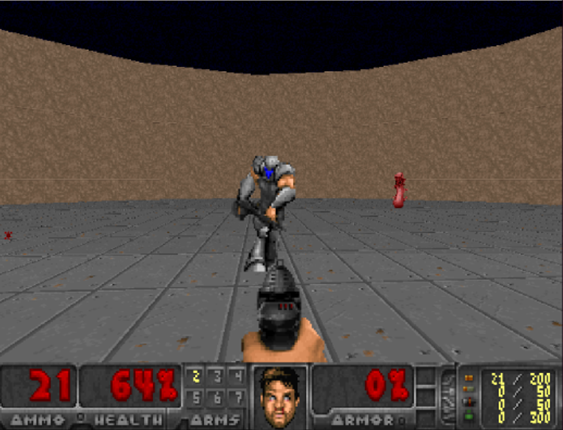}
  \caption{Defend the Center Scenario: melee enemies converge towards the center}
\end{figure}

\section{DEEP REINFORCEMENT LEARNING OVERVIEW}

\subsection{Reinforcement Learning, Q-Learning, and Deep Q-Networks}

The field of reinforcement learning frames a learning problems using two separate entities: an agent and an environment. As Sutton discusses [10], it is useful to conceptualize this framework as follows: an agent is defined as an actor attempting to learn a given task. The agent is delineated from the environment by defining its characteristics and action set as items encompassed by its realm of influence and under its complete control; all other aspect of the learning problem are then attributed to the environment. Generally, the environment can be described as the setting in which the learning problem takes place. The agent receives its current state \( s_{t} \) from the environment, and it proceeds to interact and affect the environment through action \( a_{t} \) , which is derived from a policy \( \pi_{t} \) . The environment takes in action \( a_{t} \) and updates the current state of the agent to \( s_{t+1} \) along with sending a reward signal \( r_{t} \) to the agent, indicating the value of \( s_{t} \) to \( s_{t+1} \) transition via action \( a_{t} \). Given the RL framework, we can define discounted rewards at time \(i=0\) as :

$$
R = \sum_{i = 0}^{T}\gamma^i r_i \eqno{(1)}
$$

, where \( \gamma\) \( \in\) [0,1). Rewards are discounted to simulate the concept of delayed rewards to the agent, the importance of the discounted being directed by the value of \(\gamma\). \(\gamma\) values close to 0 indicate immediate rewards are more important; whereas \(\gamma\) values close to 1 indicate to the agent that longer of sequences of actions are important to consider to achieve high rewards. The goal of the agent is to maximize its rewards as seen by the environment over the course of a single experience or series of experiences (i.e games) by developing a policy. This policy dictates what action an agent performs given its current state.

To obtain an approximation of optimal values corresponding to state-action pairs, the temporal difference method [11] Q-learning, first developed by Watkins [3] in 1989, can be used. 
The Q-function for a given state-action pair with policy \(\pi\) is as follows,

$$
Q^{\pi}(s,a) = E[R_i | s_i = s, a_i = a] \eqno{(2)}
$$

The task is then to find the optimal policy, giving:

$$
Q^{*}(s,a) = max_{\pi}(Q^{\pi}) = max_{\pi}(E[R_i | s_i = s, a_i = a]) \eqno{(3)}
$$

Thus, Bellman optimality is then satisfied by the optimal Q-function, as the above can be rewritten as:

$$
Q^{*}(s,a) = max_{\pi}(Q^{\pi}) = E[r + \gamma * max_{a'}( Q^{*}(s',a') )  | s, a ] \eqno{(4)}
$$

For most problems - with large state-action spaces - direct, recursive methods, including dynamic programming methods, are impractical. Rather, the optimal Q-function is approximated by a parametrized method. In particular, deep learning architectures have proven very successful at approximating the optimal value of high-dimensional target spaces. Let \(Q_w \) be defined as a Q-function with parameters w. For updates to the \(Q_w \) function, the loss can be defined as:

$$
L_{t}(w_{t}) = E_{s,a,r,s'}[ l_{t}(y_{t} -  Q_{w}(s,a)) ] \eqno{(5)}
$$

, where \(l_{t}\) is any reasonable transformation of the temporal difference error for training deep neural networks, including L1 (proportional to mean absolute value error) and L2 (proportional to mean squared error) losses, among others, 
and

$$
y_{t} = r + \gamma * max_{a}(Q_{w}(s',a))
$$

Given successes with stochastic gradient descent updates, we can instead lose the expectation and give stochastic updates to \(Q_{w}\) using the following as loss for backpropagation:

$$
L_{t}(w_{t}) = l_{t}(y_{t} -  Q_{w}(s,a)) \eqno{(6)}
$$

To gain experiences in the environment, \(\epsilon\)-greedy training can be used, wherein the agent randomly acts at with probability \(\epsilon\) or choses what it deems its best action with probability 1 - \(\epsilon\). Using this strategy, \(\epsilon\) is decayed over the course of training, usually starting with a value of 1 and having a minimum value of 0.1.

To stabilize Q-values during learning, a replay memory is used to remember \(s,a,r,s'\) experiences as the agent interacts with the environment; the agent then learns by sampling from this replay memory uniformly after a set number of games played and fitting against the selected experiences. The replay memory along with a target Q-function were introduced to counter the algorithm’s strong affinity to local optima given these greedy updates.

Enter the modern framework for Deep Q-Networks in full. Deep Q-Networks (DQNs) are a a parametrized, model-free, off-policy method; specifically, it uses Q-learning for value estimation. Two deep neural network architectures are used to learn the Q-values for each experience sampled from replay memory. An online network \(Q_w\) learns in a greedy fashion, and the target network \( \tilde{Q}_w \) acts as a tether point to reduce the likelihood of the online network from falling into a local optima due to its greedy updates.

Let \(Q(s,a)\) be defined as the online network, and \( \tilde{Q}(s,a) \) be defined as the target network. During training using replay memory after a number of games have been played, the online network is updated using the following target:

$$
Q_{target}(s,a) = r + \gamma max_{a}(\tilde{Q}(s',a)) \eqno{(7)}
$$

\subsection{Deep Recurrent Q-Networks}

A variant of DQN, deep recurrent Q-Networks (DRQNs) have shown exceptional performance in a variety of RL problems. In environments characterized by significant latent information, recurrency offers a way to fill in the gaps of missing knowledge for a given state[7]. The RL environment of ViZDoom is no exception; as a first person shooter, the agent is bound by a first-person, 90-degree view of objects in front of it. There is no radar - a HUD display of enemies around the player in a 360 degree arc is present in many other FPS games - or other indicators of what could be in the other 270 degrees.

As such, rather than receiving states \( s_t \) as in the RL framework, it is more aptly put that the agent in ViZDoom receives partial observations \( o_t \) of its current state \( s_t \) [7]. Thus, using a DQN, the objective is to estimate not \( Q(s_t, a_t) \) but instead \( Q(o_t, a_t) \), putting the agent at a distinct disadvantage concerning state level information. One way to counter this is to include state information from the previous sequence of states, thereby allowing the agent to instead estimate \( Q(o_t, h_{t-1}, a_t) \) where \(h_{t-1} \) represents information gathered at state \(t-1\) and passed to state \(t\). The long-term, short-term memory cell, LSTM, is one recurrent construct capable of doing this; at a given time \(t\), the LSTM cell takes in \(o_t\) and \(h_{t-1}\) and outputs \(h_t\). Instead of \(Q(o_t, a_t)\), the network then estimates \(Q(h_t, a_t)\), increasing the level of information available to the agent for a given state-action pair.

\subsection{Double-Q Learning}

Double-Q learning advances upon DQN in a simple, yet remarkably effective way: let the loss used in learning by the agent be defined by the value-maximizing actions of the online network with the Q-values of the target network associated with those maximizing actions. With the decoupling of the maximizing action from its value, it is possible to eliminate the maximization bias present in Q-Learning [6]. Specifically, the loss is defined as follows:

$$
L_t(w_t) = l_t(y_t -  Q_{w}(s,a)) \eqno{(8)}
$$

,where \(l_t\) is a transformation of the temporal difference (TD) error and

$$
y_t = r + \gamma \tilde{Q}_{w}(s’, a_{online}))
$$

$$
a_{online} = argmax_{a}(Q(s’, a)) 
$$

This innovation was spurred by the observation that Q-learning tends to overestimate the values of state-actions pairs. Through experimentation of deep double-Q learning using myriad, diverse Atari game environments, Hasselt et. al. find that deep double Q-networks(DDQNs) show marked improvement in the estimation of values for state-action pairs [14]. Furthermore, Double-Q learning introduces a level of stability in Q-learning updates, which allows for more complex behavior to be learned.

\subsection{Prioritized Experience Replay}
During training, the DQN agent samples uniformly from its replay memory to populate a batch training set and subsequently learn on this set. This is done many times over the course of the experiment, allowing the agent to continue to learn from previous experiences. The construct of replay memory was created to simulate learning from experiences that are sampled i.i.d (independently and identically distributed). Without approximating i.i.d sampling, the agent can quickly overfit to recent state-action pairs, inhibiting learning and adaptation. The method of prioritized experience replay innovates on this front by biasing the sampling of experiences [12]. Specifically, experiences are weighted and sampled with higher probability according to the TD error observed for that sample - the larger the TD error, the higher the probability with which a given experience will be sampled. The intuition behind this is that experiences that significantly differ from the agent’s expectation will have more didactic potential. Concerning empirical validation of the method, Schaul et al. created PER and showed that it affords significant performance improvement across many Atari-based RL environments. The authors here apply PER to a DRQN model and train and test it in the defend-the-center ViZDoom scenario (see IV. A. Scenario section). 

\subsection{Ensembles of Deep Q-Networks - Snapshot Ensembles}

Ensembling of models has proven useful in many situations, countering, to an extent, the tendency of nonlinear models to over-fit to a given training distribution. However, the generation of ensembles for deep neural networks can prove onerous, requiring the training of multiple networks in parallel or - even worse, in terms of total training time - sequentially. Recently, methods have been proposed to gain the advantages of ensembling while reducing the time to generate such a set of models. Snapshot ensembling is one such example, employing a cosine annealing learning rate to generate \(M\) models over \(T\) total epochs from the training of a single model over those \(T\) epochs [4]. This is done by using the cosine annealing learning rate to train the single model and take “snapshots” of its current weights every \( T/M \) epochs. Thus, only a single model is trained while, at the same time, providing a diverse model population for the ensemble through use of the cosine annealing learning rate.  

The authors here use the snapshot ensemble method to analyze performance improvement of the aforementioned model (DRQN) in the context of the ViZDoom Reinforcement Learning Environment, utilizing the learning enhancement methods of PER and DDQ.

\subsection{Review of Modern RL Game Environments}

\subsection*{For Reinforcement Learning, Why Video Games?}

A burgeoning field of work has been created using the VizDoom environment due to the unique problems that the 3-D first-person shooter (FPS) engine can provide.  Navigation, recognition of objects in the space, and decision making when other objects or actors are encountered are all obstacles that an FPS environment presents.  Another important obstacle is that the agent will never have a complete view of the given state space.  Recent deep reinforcement learning methods applied to this environment have focused primarily on the free-for-all death match setting to train their models.  Instead, the authors here choose to train and test models using the Defend the Center mode.

Including ViZDoom, there are a number of diverse game-based RL frameworks in current use. In general, video games have increasingly become an important tool for AI research and development. It is easy to see why, as video games allow for an environment rich with parameters, feedback, and end goals for agents to gauge their success.  Originally, the frameworks tested were simple 2-D environments from games originally on the Atari 2600 that allowed for simple movement and near fully observable states. On the other hand, the engines for FPS environments provide a wealth of data and multiple obstacles and objectives for the AI agents to learn from. One of the most important obstacles provided by the FPS setup is the lack of complete information for a given state. This obstacle mimics real world problems for autonomous agents, as they must be able to act with a limited view of the world around them.

\subsection*{Arcade Learning Environments}

The classic 2D video games of the past have been used to a great extent for deep learning. The main platforms for these games are the Atari 2600, Nintendo NES, Commodore 64 and ZX Spectrum. One of the most used emulator environments is Stella, which uses the Atari 2600 and has 50 games available. Methods previously used in these environments include Double DQN, Bootstrapped DQN, Dueling DQN, and Prioritized DQN. Montezuma’s Revenge is a notable game from this genre as it requires memorization of other rooms that aren’t immediately available to the agent. Methods used in this space are DQN-PixelCNN, DQN-CTS, and H-DQN [2, 9]. 

\subsection*{Real Time Strategy}

StarCraft: Brood War is a popular real time strategy (RTS) game in which the objective is to destroy all of the enemy structures and claim victory. Players move towards victory by gathering resources to create attacking units that can then be directed in various actions. The obstacles for the agent are many as the state space is complex and not fully observable by the agent at any one time. There are three factions available in this environment that all have their own unique characters and abilities. Even if the agent is limited to learning to play only a single faction, there are still 3 different match ups that could be encountered. Each of these match ups will have their own strategies and units that will be needed to counter different compositions of units built from different structures. 

Another important skill that the agent needs to learn is intelligence gathering - understanding of both map layout and composition of the opponent's army. At the start of the game, the map is under what is called “fog of war”, which blacks out any area that doesn’t have a controlled unit to provide vision. In summary, the agent must navigate the environment and gather intelligence on the opposing player, build an appropriate base from which to train units to defend and attack the opponent, maneuver these units into advantageous positions in the environment, engage opponent units and manage the abilities available to units, and manage resources. All of these tasks must also be performed without complete knowledge of the state space. The combination of the many problems experienced in the course of one game has led deep learning researchers to focus on specific problems in the game as the sparsity of rewards makes the training of highly non-linear functions, such as neural networks, difficult. The main problem that has been the focus of many researchers has been the micromanagement of units in combat scenarios. Methods such as  IQL, COMA, Zero Order, and BiCNet have been performed with promising results [9]. 

\subsection*{Open World Games}

Open world games are another avenue of research, as the nature of open world games positions issues of exploration and objective setting as the main challenges to the agent. Project Malmo is an overlay built onto the Minecraft engine that can be used to define the environment and allow for objectives to be imposed on an agent in a normally free task environment.  These large and open problems are commonly used to test reinforcement learning methods. Methods such as H-DLRN and variations of NTMs have been successful in preforming varying tasks in the space [9].

\subsection*{Racing Games}

Racing games are another popular genre for AI research as there are many challenges here as well. Depending on the game chosen, the inputs for control can be as complex as having a gear stick, clutch and handbrakes, while others are much more simplified. Challenges for this environment include positioning of the vehicle on the course for optimal distance traveled, adversarial actions to block or impede other drivers when other vehicles are present, and sometimes the management of resources. This genre is also useful because the entire state is not available to the agent in the first or third person view. Methods that have been used in this genre include Direct Perception, Deep DPG and A3C, with a popular environment for this genre being the simulator TORCS [9].

\begin{figure*}[t!]
  \centering
  \includegraphics[width=0.8\textwidth]{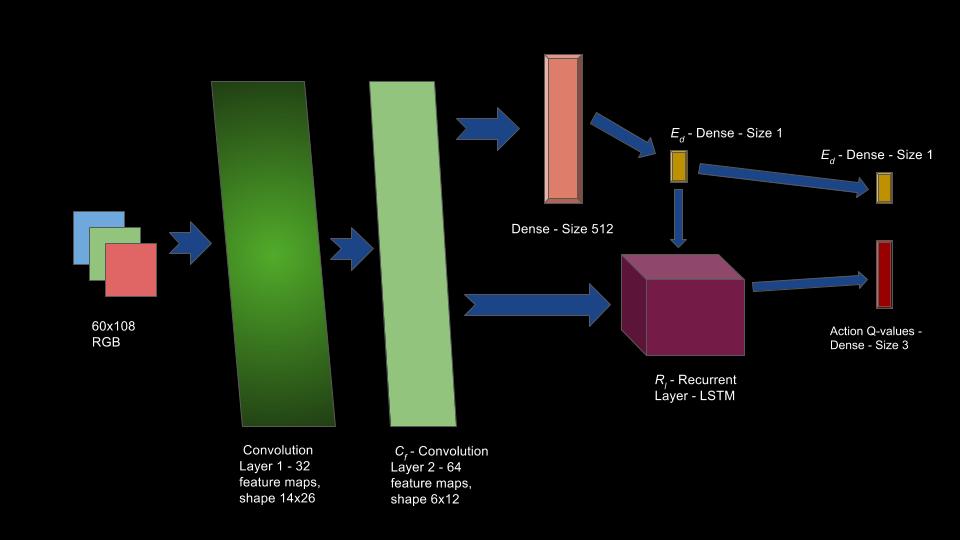}
  \caption{DRQN Architecture}
\end{figure*}

\subsection*{First-Person Shooter}

VizDoom is based off of the popular video game Doom and is a first person shooter. In this environment there are many challenges that make it a useful tool for training AI. There are many modes of play within the VizDoom environment, two of which are mentioned here: Defend the Center and Deathmatch.  In Defend the Center, the agent is spawned in the center of a circular room and is limited to turning left or right to find and eliminate the enemies that spawn. At most, only 5 melee enemies are present on the map at any one time. These are spawned against the wall; they then make their way towards the agent in the center. As enemies are eliminated, others are spawned to take their place. The objective is to hold out for as long as possible with a limited amount of ammunition. Having limited ammo helps to constrain episodes to a limited time limit, as when ammo is depleted, defeat is inevitable. In Deathmatch the objective is to reach a certain amount of frags (i.e. kills) before other actors on the map or to have the most frags when time runs out. The view of the agent is determined by the resolution of the screen chosen and varies between 90 and 110 degrees, meaning that the agent doesn’t have complete knowledge of its full state at any time. Other obstacles include the recognition of enemies in the space, aiming weapons, and navigation of space. Methods that have been used in this environment include DQN+SLAM, DFP, DRQN+Auxiliary Learning, and A3C+Curriculum Learning [8, 9].


\section{MODEL ARCHITECTURE AND AGENT TRAINING}

\subsection{Deep Neural Network Architecture}

The authors here use the DRQN model architecture (Figure 2) specified by Lample et al. [5]. This model architecture was used in all experiments, as it was noteworthy in its performance without the additions of PER and DDQ; thus the authors here endeavored to use PER and DDQ to observe how such variants can aid a deep learning architecture that is well suited to this RL problem. A general description of the architecture is as follows:
\begin{itemize}
  \item input to the network is a single frame, i.e. the 3 channels (RGB) of a frame, with each frame being resized to 60x108 pixels
  \item input is sent through two convolutional layers
  \item the output of the final convolutional layer \(C_f\) is flattened and sent to two destinations
  \item output of \(C_f\) is sent to a dense layer and then to a subsequent dense layer \(E_d\) with sigmoid activation of size 1. This is sent to a recurrent layer \(R_l\) and is also used for loss
  \item output of \(C_f\) is flattened and then sent directly to the recurrent layer \(R_l\)
  \item \(R_l\) then outputs to a dense layer with ReLu activation and then to a subsequent dense layer with linear activation of size three, as there are three actions that can be performed at a given time (turn left, turn right, and shoot) in the Defend-the-Center scenario.
\end{itemize}

The size-1 dense layer is a boolean indicator for enemy detection and is fitted by querying the game engine for enemies in the agent’s vision; if there are any enemies in the agent’s view, the true value is 1, otherwise it is 0. The feeding of predicted enemy detection information into the recurrent layer was found by Lample et al. to be very beneficial in training in a related ViZDoom RL scenario known as Deathmatch, where agents engage in a competitive, FPS match. 

\subsection{Frame Skipping, Fitting, and Reward Structure}

Frame skipping is quite crucial in the training of RL agents that use video data with a reasonably fast frames-per-second (fps) value [10]. For example, using a frame-skip value of 2, the agent will make an action on frame 0, \(f_0\). That same action will be performed for the skipped frames, \(f_1\) and \(f_2\). There are many reasons for using frame-skipping, one of the main benefits being that it prevents the agent from populating the replay memory with experiences that differ almost imperceptibly from one another. Without frame-skipping, this can pose a serious issue for training DRQN’s, as the elements of sequences that are sampled for training will be practically the same, often causing the agent to learn degenerate, loop-like behavior where it performs a single action over and over. For all experiments, the ViZDoom engine was run at 35 fps. The authors here experimented initially with various frame skip values, opting for a frame-skip value of approximately 10 for reported results.

For training of a DRQN, sequential updates are important in order to take advantage of the recurrent layer. Here, the authors sampled sequences of length 7, used the first four experiences as primer for training - passing LSTM cell state information sequentially from the previous experience to the next - and trained on the final 3 experiences.

Agents were trained for a total of 11500 games in the defend-the-center scenario, using \( \epsilon \)-greedy training. Here, \( \epsilon \)  was decayed from 1.0 to 0.1 over the course of training. For the three experiments without snapshot ensembling, a linearly decaying, cosine annealed (small amplitude) learning rate was used. For the other three using snapshot ensembling, a cosine annealed learning rate was used. Instead of using Adam, which combines classical momentum with the norm-based benefits of RMSProp, all models were optimized using the Nesterov Adam optimizer, as this swaps out classical momentum for an improved momentum formulation, Nesterov’s accelerated gradient (NAG) [13]. In terms of reward structure, the agent was given a reward during training of \(+1\) for frags of enemy units, and a penalty of \(-0.5\) for deaths. A penalty proportional to the amount of health lost was also included. 

\subsection{Hyperparameters}

Hyperparameter values for all experiments are as follows:
\begin{itemize}
  \item \(\gamma\) value of 0.9
  \item for PER, a \(\beta_{0}\) value of 0.5 was used, and \(\beta\) was increased to 1.0 linearly over the course of each experiment
  \item for PER, \(\alpha\) value of 0.7
  \item for PER, \(\epsilon\) offset of 0.05 for non-singular priority values
  \item a batch size of 20 with sequence length 7 was used for replay memory training
\end{itemize}

\section{EXPERIMENTS}

\begin{figure}[b!]
  \centering
  \includegraphics[width=0.5\textwidth]{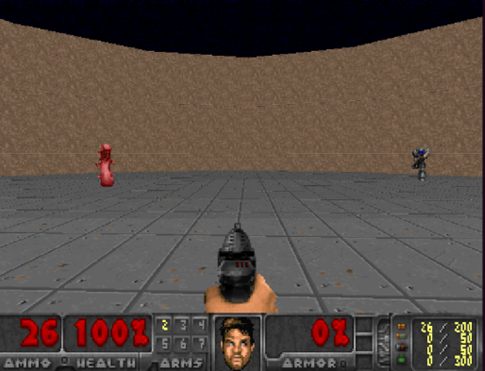}
  \caption{Defend the Center Scenario: enemies spawn at a distance from the agent}
\end{figure}

\subsection{Scenario}

Previous literature concerning ViZDoom has focused on using the Deathmatch scenario to train their agents in navigation, combat, and enemy recognition. The authors here focus on the problem of spatial recognition as opposed to navigation. In order to emphasize this, the game mode Defend the Center available in the Vizdoom engine was chosen. This goal of this game type is to frag as many adversaries as possible before you are overrun. The agent is allowed no movement in the center of this circular arena other than the adjustment of its angle of view.  With this limitation, the authors here aim to have agents learn spatial awareness and to prioritize the targets based on distance from the agent.  As a review of Section II F, in the Defend the Center scenario, the agent is spawned in the center of a circular room and is limited to altering the degree of its view to find and eliminate the enemies that spawn. At most only 5 melee enemies are spawned against the wall (Figure 3) that will then make their way towards the agent in the center. The objective is to hold out for as long as possible with a limited amount of ammunition.

\subsection{Software and Hardware}

Six experiments in total were performed, two sets of 3 experiments each. The first set was to test the base DRQN and PER and DDQ variants without snapshot ensembling, while the second set employed the use of snapshot ensembling. In both cases, the authors here use the proportional PER version, rather than rank PER. In all experiments, agents were trained for approximately 12 hours for a total of 11500 games of the ViZDoom defend-the-center scenario on an NVIDIA GeForce Titan X GPU using Python 3.7 and the neural network library Keras, with Tensorflow backend and the NVIDIA CUDA Deep Neural Network library (cuDNN).  





\begin{figure}	
	\centering
	\begin{subfigure}[b]{4in}
		\includegraphics[width=0.9\textwidth]{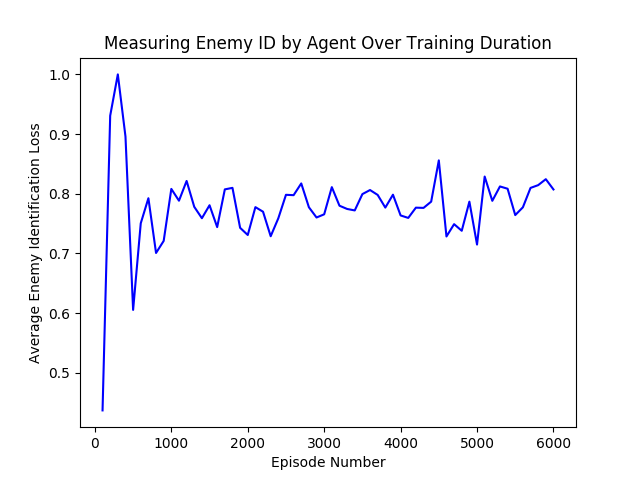}
		\caption{DRQN}		
	\end{subfigure}
	\begin{subfigure}[b]{4in}
		\includegraphics[width=0.9\textwidth]{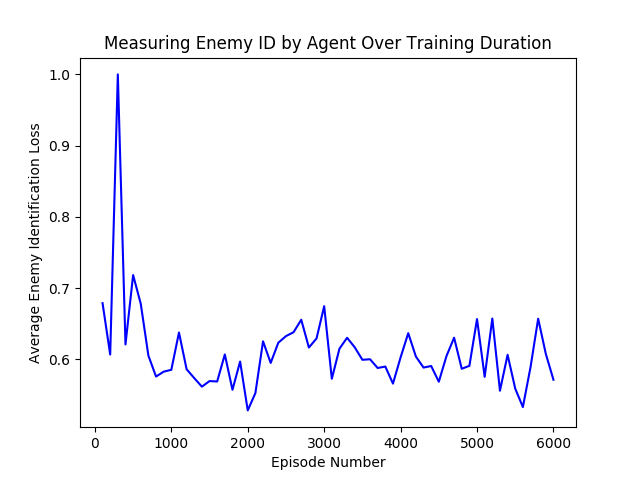}
		\caption{DDQ DRQN}
	\end{subfigure}
    \begin{subfigure}[b]{4in}
		\includegraphics[width=0.9\textwidth]{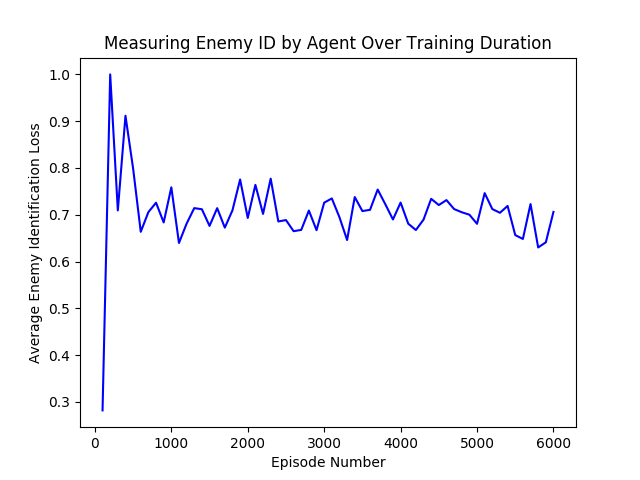}
		\caption{PER DRQN}
	\end{subfigure}
	\caption{Average Enemy ID Cross Entropy Loss - Normalized by Max Loss}
\end{figure}


\subsection{Results}

The aim of these experiments is to test the learning benefits offered by DDQ and PER - relative to the baseline DRQN architecture that was formulated by Lample et al. and reproduced here - in the context of ViZDoom, an environment where a great deal of state information is hidden from the agent at each time step. Specifically, the authors here studied effects of DDQ and PER on the early phases of learning, using a sizeable frame-skip value. As stated above, a frame-skip value of approximately 10 was used for training of all agents. Given that most research using ViZDoom has studied RL agents using much smaller values of frame-skip (around 5), this work allows a look at RL agent learning rate improvements at higher frame-skip values in the context of this FPS environment. As noted by Braylan et al., using larger frame-skip values greatly accelerates learning [1]. Such behavior was also noted by authors here; larger frame-skips led to significant accelerations in learning rate, albeit at the cost of the agent’s precision. With larger frame-skip values, the agent can overshoot an enemy when turning toward the target, decreasing the ability of the agent to center on target before shooting. At the other end of the spectrum, too few frame-skips can cause the agent to learn degenerate, repetitive behavior - as noted earlier in this paper - such as assigning higher value, irrespective of input, to the action with the highest variance in value.

For testing, snapshots of each agent - base DRQN, DRQN with DDQ, and DRQN with PER - were taken at 100 game intervals over the course of the full 11500 games of training. These snapshots were then tested in defend-the-center games and, as with training, tasked with gaining as many frags as possible. Specifically, each snapshot was given 100 games to accumulate frags. In addition, three sets of five models each were used in creating snapshot ensembles - one ensemble each for the base DRQN, DRQN with DDQ, and DRQN with PER agent types.

Given the higher frame-skip used, the authors, unsurprisingly, observed that in all cases the performance of agents plateaued after roughly 6000 out of the 11500 games. Given that the subject of interest here is the rate of learning in early development of the agent, this is not an issue. However, it does serve as a basis for further research. In particular, one might ask what the upper limit of performance would be at smaller frame-skip values using the aforementioned DRQN architecture with the addition of either PER or DDQ. This is one of a set of future targets for further investigation. 

Concerning metrics of performance, cross entropy loss and K/D ratio are used to measure agents success at various time points in development. Cross entropy loss was used to measure the agent's recognition of an enemy or enemies in the current input frame, calculated using the output of the size-1 dense layer (see Figure 2). K/D ratio is the frags-to-deaths ratio, i.e. the number of total frags by the agent over its total number of lives. Roughly speaking, this translates to frags-to-games-played ratio.


\begin{figure}	
	\centering
	\begin{subfigure}[b]{4in}
		\includegraphics[width=0.9\textwidth]{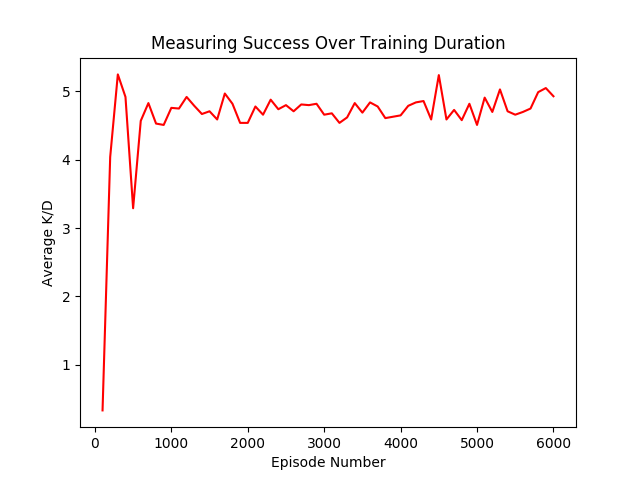}
		\caption{DRQN}		
	\end{subfigure}
	\begin{subfigure}[b]{4in}
		\includegraphics[width=0.9\textwidth]{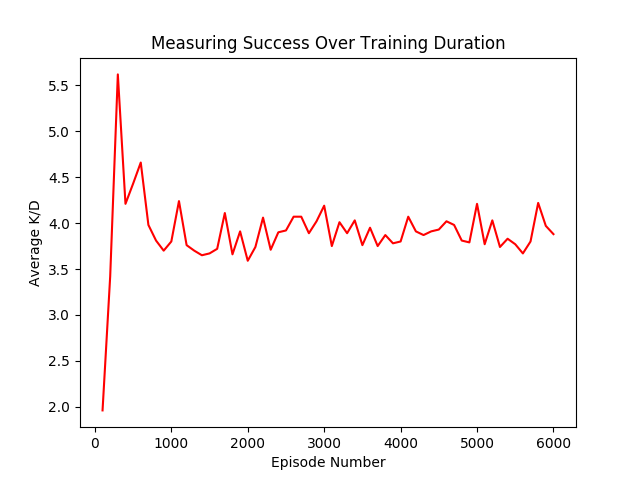}
		\caption{DDQ DRQN}
	\end{subfigure}
    \begin{subfigure}[b]{4in}
		\includegraphics[width=0.9\textwidth]{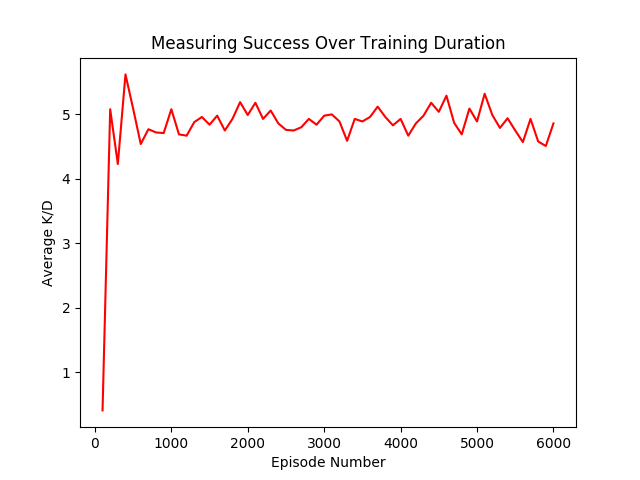}
		\caption{PER DRQN}
	\end{subfigure}
	\caption{Average K/D over 100 Defend the Center Games}
\end{figure}

In the case of DRQN with PER, the agent quickly achieved impressive levels of performance, reaching an average K/D of 5.62 after 400 games trained (see Figure 5). The maximum average value (averaged over 100 games for each model and then finding max over all models) for enemy identification cross entropy loss was 0.2196 (see Figure 4 - all loss was normalized by the max loss value observed for each graph), with a mean average K/D of 4.82 and an average K/D standard deviation of 0.614. The snapshot ensemble method improved upon the mean average K/D of DRQN with PER by over a standard deviation, achieving an average K/D of 5.51. Likely, the snapshot ensemble of PER was so successful due to PER’s aggressive tendency to fit against challenging samples mixed with the snapshot ensemble method’s ability to create similar and yet competitively diverse committees using its cosine annealing learning rate.

\begin{table}[h!]
\caption{K/D and Enemy ID Loss}
\begin{center}
\begin{tabular}{|c||c||c||c|}
\hline
& DRQN & DDQ DRQN & PER DRQN\\
\hline
Mean Average K/D & 4.65 & 3.90 & 4.82\\
\hline
Average K/D Standard Dev. & 0.620 & 0.393 & 0.614\\
\hline
Max Enemy ID Loss & 0.269 & 0.298 & 0.2196\\
\hline
\end{tabular}
\end{center}
\end{table}

For the DRQN with DDQ, the authors report a set of agents characterized by consistency. Cumulatively, these agents had a mean average K/D of 3.90 and an average K/D standard deviation of 0.393, and the maximum average value for enemy identification cross entropy loss was 0.298. The snapshot ensemble method improved upon DRQN with DDQ by over a standard deviation as well, achieving an average K/D of 4.37 .

Using the DRQN architecture mentioned here (without PER or DDQ), the agent achieved a max average K/D of 5.25, never reaching an average K/D to rival the max average K/D observed by the agent with PER. Furthermore, the base DRQN agent had an average K/D standard deviation of nearly double that of the agent with DDQ, illustrating its lack of consistency relative to the DDQ version of itself. To elaborate, the base DRQN agent had a mean average K/D of 4.65, an average K/D standard deviation of 0.620, and the maximum average value for enemy identification cross entropy loss was 0.269. No improvement was observed using the snapshot ensemble of base DRQN’s, indicating potentially that increased value overestimation in such an environment led to development of a committee of models that were too dissimilar to allow for advantages offered by ensembling. Note that the DRQN architecture used here is efficient and very competitive, allowing for quick learning of complex tasks. Thus, PER and DDQ should be looked on as boosts to a well tuned learning architecture.

In all experiments, successful enemy detection was achieved by 1000 games by all agent types. In terms of qualitative notes on agent behavior, DDQ DRQN agents were better at developing and maintaining an ammo conservation strategy. Specifically, it was noted that DDQ agents would hold fire longer than other agents, allowing enemies to come closer. This in turn, reduced the possibility of missing a shot. DRQN PER agents, on the other hand, were observed to more quickly respond to enemies attacking from behind. When the agent is damaged, its screen will flash red, and the amount of health remaining is indicated in the lower left of the screen. PER agents were seen to more quickly turn to address and neutralize attackers from behind as well as from the flanks.

\section{CONCLUSION}

In this paper, the authors tested double-Q learning and prioritized experience replay in the context of the ViZDoom reinforcement learning environment. In addition, these methods were coupled with the use of the efficient deep neural network ensemble creation method known as snapshot ensembling. This breaks new ground in the ViZDoom environment, as DDQ and PER had not been tested using this clever, ensembling method - populating members of an ensemble by effectively training a single model, in this case a deep recurrent neural network, with the use of a cosine annealing learning rate. Furthermore, snapshot ensembling yielded significant results, improving DDQ and PER agent performance by over a standard deviation above the mean in both cases.

Following this, there are a number of avenues for further research. One of these consists of examining other DRL models at higher frame-skip values using ViZDoom. To what extent are other DRL model structures (dueling network architectures, actor-critic models, asynchronous variants, etc.) affected? Not only does frame-skipping accelerate training, it can also be looked on as modeling a potential scenario for autonomous agents in physical 3D space. If on a mission for example, a search-and-rescue agent's visual sensors are damaged and only every 11th frame on average of its video feed is a valid image, a reasonable question that arises is: will the agent, trained on a frame-skip distribution with a mean of \( x \) be able to cope with the new, shifted frame-skip distribution with mean \(x+k\), where \(k\) is the additional number of frames, on average, that must be skipped before a valid frame is observed? At what values of \(k\) does performance begin to degrade significantly. This is also assuming the distribution shape remains unchanged; what are the limitations of learning from a certain type of frame-skip distribution (uniform, Gaussian, Landau, etc.)?  Given its emerging research community and solid RL environment, the authors here note ViZDoom as a promising setting to test further research questions such as these.

\addtolength{\textheight}{-12cm}   




\section*{ACKNOWLEDGEMENT}

The authors here would like to thank: 1) the ViZDoom development team for their continued maintenance and extension of this remarkable RL framework, and 2) IEEE CIG for their support of the annual ViZDoom Limited and Full Deathmatch Competitions.


\end{document}